\documentclass[11pt]{article}

\usepackage[final]{acl}

\usepackage{times}
\usepackage{latexsym}

\usepackage[T1]{fontenc}

\usepackage[utf8]{inputenc}

\usepackage{microtype}

\usepackage{inconsolata}

\usepackage{graphicx}

\usepackage{booktabs}
\usepackage{multirow}
\usepackage{amsmath}
\usepackage{cleveref}

\title{Am I More Pointwise or Pairwise? \\Revealing Position Bias in Rubric-Based LLM-as-a-Judge}

\author{
  Yuzheng Xu\textsuperscript{1,2} \quad
  Tosho Hirasawa\textsuperscript{1,2} \quad
  Tadashi Kozuno\textsuperscript{1,2} \quad
  Yoshitaka Ushiku\textsuperscript{1,2} \\
  \textsuperscript{1}OMRON SINIC X \quad
  \textsuperscript{2}NexaScience \\
}

\begin{document}
\maketitle
\begin{abstract}
Large language models are widely employed as evaluators, a paradigm commonly referred to as LLM-as-a-Judge. Prior research has predominantly examined point-wise or pair-wise evaluation protocols; in contrast, our focus is on rubric-based evaluation, which has been attracting increasing attention owing to its utility for training models in domains where verification is otherwise difficult. In this work, we show that rubric-based evaluation implicitly resembles a multiple-choice setting and therefore exhibits position bias: LLMs tend to prefer score options that appear at specific positions within the rubric list. Through controlled experiments across multiple models and datasets, we demonstrate that this position bias is consistent. Its direction, however, is model-specific: some judges favor the first option, while others favor the last. We further identify a second, orthogonal axis of bias: when a prompt scores several criteria simultaneously, the ordering of the criteria itself shifts the resulting scores. We additionally explore permuting the order of the rubric options as a means of mitigating position bias, and find that although the bias can be attenuated, improvements in the correlation between model judgments and human annotations are obtained primarily for models that exhibit strong bias. Our results recast rubric-based LLM-as-a-Judge as a multiple-choice problem with measurable, model-specific position bias, and we further confirm that only a small number of random order permutations are sufficient to reduce the error introduced by this bias for the majority of models.
\end{abstract}

\section{Introduction}

Large language models (LLMs), exemplified by ChatGPT~\cite{openai2023gpt04}, have been widely applied to language-related tasks due to their strong performance, such as summarization~\cite{goyal2022news, bhaskar-etal-2023-prompted}, question answering~\cite{rein2024gpqa}, and even creative idea generation~\cite{lu2024ai, xu2025mk2}. 
Due to their strong generality on text-related tasks, LLMs can also be used to generate evaluations of textual quality, a paradigm commonly referred to as LLM-as-a-Judge~\cite{mtbench, ye2024justiceprejudicequantifyingbiases, gptscore}. 
This capability is useful not only for producing evaluation scores but also for training LLMs themselves~\cite{rlaif}, leading to a wide range of applications.
Although LLMs are highly capable in this regard, their performance has also been questioned. Some prior work has pointed out that using LLMs as judges for text quality evaluation can introduce various biases like position bias, verbosity bias, and so on~\cite{mtbench, ye2024justiceprejudicequantifyingbiases, chen-etal-2024-humans, huang-etal-2025-empirical}. This phenomenon needs to be carefully considered in practice. 

\begin{figure}[t!]
\centering
\includegraphics[scale=0.63]{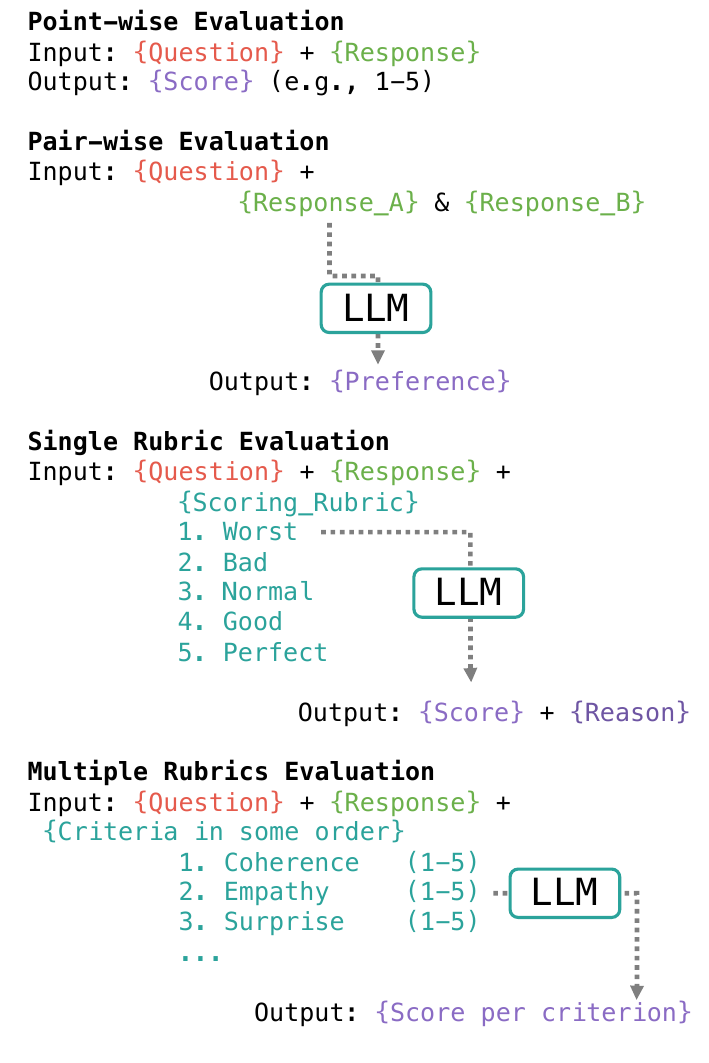}
\caption{Four paradigms of LLM-as-a-Judge evaluation. Point-wise evaluation assigns a score given a question and a single response. Pair-wise evaluation compares two responses and outputs the model’s preference. Rubric-based evaluation further incorporates explicit scoring criteria. We can explicitly display the rubric scores along with their corresponding descriptions, or we can simply specify the parts that need to be evaluated and the specific score ranges. Judge models may also exhibit position bias toward responses in certain orders. }
\label{fig:pevals}
\end{figure}

\begin{figure*}[t!]
\centering
\includegraphics[scale=0.4]{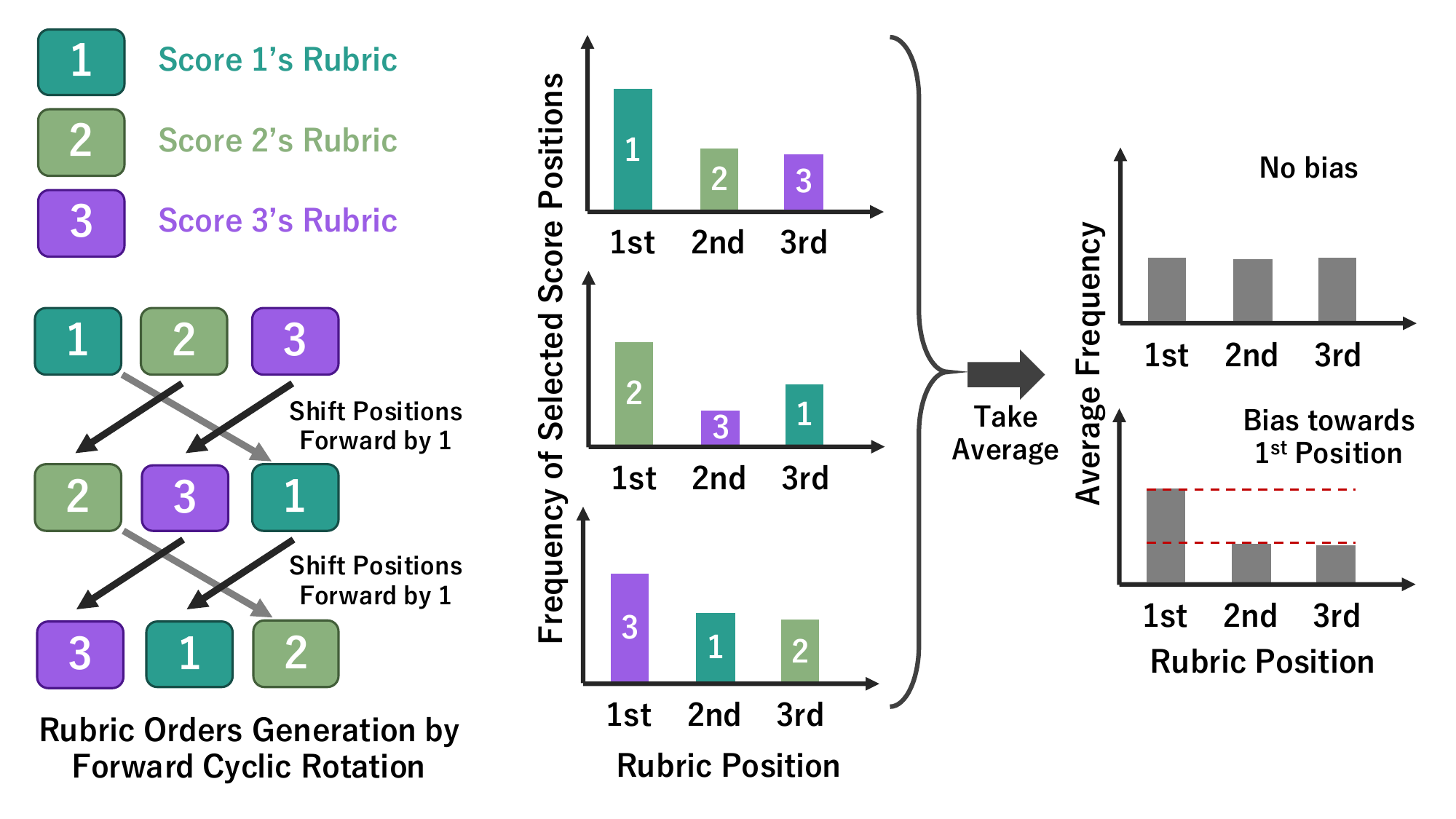}
\caption{Balanced permutation of rubric orderings. Aggregating the model’s choice distributions across permutations marginalizes out score identities and reveals systematic position bias.}
\label{fig:pteaser}
\end{figure*}

The presence of bias can also depend on the evaluation type. As shown in~\Cref{fig:pevals}, in practice, two popular evaluation paradigms are commonly used: point-wise and pair-wise~\cite{tripathi2025pairwise}. The former assigns a score to a given response, while the latter selects the better one between two candidate responses. Constructing rubrics and asking the model to perform evaluation based on them is also a widely adopted approach~\cite{kim2024prometheus}, as it provides greater flexibility. We can write out in detail the score and description corresponding to each rubric, or we can directly list all the dimensions to be evaluated, omit the details, and simply describe the score range.

In this paper, we focus on the rubric-based LLM-as-a-Judge setting. Since there is only a single text to be evaluated, there should be no position bias in the classical sense. However, the rubric-based LLM-as-a-Judge can be viewed as a multiple-choice selection problem: given a text for evaluation, the LLM must choose the rubric option that aligns most with the text.
Unlike the pair-wise setting, positional bias in rubric-based evaluation is less salient, because score selections are inherently imbalanced and dominated by the underlying score distribution, making positional effects harder to observe directly. Additionally, if we put all the rubrics together (the fourth paradigm), although the multiple-choice situation disappears, would the order of the different aspects affect the final output?

In our work, inspired by position bias observed in pair-wise evaluation, we propose a balanced permutation method to disentangle the influence of score values from positional choice preferences in rubric-based LLM-as-a-Judge. By ensuring that each score is evenly distributed across different positions over multiple evaluations, our method enables us to uncover position bias in rubric-based evaluations.~\Cref{fig:pteaser} illustrates how aggregating model choices across balanced rubric permutations reveals systematic positional preference. While debiasing does not always improve correlation with human scores, the ordering effect is practically consequential: it changes downstream rankings and top-1 selections even when average score correlations remain similar.

Our contributions are:

\begin{itemize}

\item We frame rubric-based LLM-as-a-Judge as an implicit multiple-choice task and show that it carries statistically significant, model-specific position bias across six judges and four datasets. Some judges are first-biased while others are last-biased, contradicting the assumption of a uniform first-option preference.

\item We identify a second, orthogonal axis of bias, namely the order of criteria within a multi-criterion prompt, which proves significant.

\item Using budget-matched ablations and a paired bootstrap, we find that balanced permutation is statistically indistinguishable from random aggregation and improves human correlation only for strongly biased judges, so a small number of randomly ordered variants suffices.

\item We show that the bias propagates downstream: rubric ordering alone flips the top-1 candidate on 15\% to 41\% of prompts, with direct implications for best-of-$N$ and rubric-as-reward pipelines.

\end{itemize}

\section{Related Work}

\paragraph{LLM-as-a-Judge: }
Human evaluation of text is an expensive option. With the advent of RLHF~\cite{rlhf}, LLMs have acquired a certain degree of alignment with human values, which makes LLM-as-a-Judge feasible~\cite{mtbench}. ~\citet{gptscore} first explored the use of GPT models for evaluating text data. This work addressed a long-standing limitation of traditional evaluation: it's hard to flexibly and easily adapt evaluation criteria to different tasks and requirements. \citet{mtbench} systematically studied LLM-as-a-Judge by introducing MT-Bench. Their results show that strong models used as LLM-as-a-Judge can closely match human preferences, achieving over 80\% agreement. Such a high level of agreement enables LLMs to be used even for training other LLMs, achieving performance comparable to RLHF. This approach is known as RLAIF~\cite{rlaif}. Along with RLAIF, some work uses rubrics as training signals~\cite{rubricsasrewards, dineen-etal-2025-qa}. Rubrics can also be used to evaluate the quality of CoT (Chain-of-Thought)~\cite{sheng2026reinforcingchainofthoughtreasoningselfevolving} or the answers without ground truth.

Our work is positioned within the line of research on LLM-as-a-Judge and focuses on a previously underexplored issue in rubric-based evaluation. We identify position bias induced by rubric ordering and propose a simple, model-agnostic mitigation strategy.

\paragraph{Evaluation Biases: }
Similar to humans~\cite{chen-etal-2024-humans}, LLM-as-a-Judge also exhibits biases. A notable example is position bias: in a pair-wise evaluation setting, the model may tend to prefer either the first or the second response, depending on factors such as the model’s capability and the quality of the answers. This bias can be mitigated by swapping the order and performing evaluation twice~\cite{mtbench}. LLMs also show length bias, tending to assign higher scores to longer, more verbose responses~\cite{mtbench, tripathi2025pairwise}. ~\citet{shi-etal-2025-judging} argue that the quality gap between candidate responses has a significant impact on position bias, while length difference is not. In addition, an LLM may favor outputs generated by itself or by closely related models~\cite{chenbeyond, chen2025llm, mtbench}. \citet{wataoka2024selfpreference} explain this phenomenon by noting that LLMs tend to prefer responses with lower perplexity, and outputs generated by the same model or closely related models often have relatively lower perplexity. Related to our work, \citet{li2025evaluating} shows that different rubric orderings can affect the correlation between LLMs and human scores. We extend this line of work by formulating rubric ordering effects as measurable position bias and by introducing a balanced permutation protocol for auditing and quantifying it. Besides, we identify an orthogonal axis, the order of criteria in a multi-criterion prompt, which is position-biased even when no explicit score-option list is presented. Our results further show that, under a matched inference budget, simple randomization often matches exact balancing, yielding a practical low-cost recommendation.

\section{Problem Formulation and Methods} 
Our work aims to study whether LLMs suffer from position bias in the rubric-based format. We have two settings in total. In the first setting, an LLM is asked to judge a question-response pair with a set of rubric score options. We define position bias in this context as a systematic preference for selecting score options appearing at particular positions. In the second setting, each rubric does not come with a specific score description; instead, the model assigns a score within a given range based on its own understanding. Please refer to~\Cref{fig:pevals}.

\subsection{Balanced Permutation}
\label{sec:balanced_permutation}
In studies of position bias in pair-wise evaluation, a common mitigation is to swap the two responses to cancel out the effect. In the rubric-based setting, however, we have multiple score options, which makes a simple swap ineffective. Nevertheless, as long as we can ensure that each score appears equally often at each position, we can in principle factor out the influence introduced by the score values themselves.

To address this, we propose a balanced permutation method. The key insight is that position bias can be neutralized if each score option appears in each position an equal number of times across multiple evaluation runs. For example, if there are 5 scores, we construct 10 complementary orderings: 5 forward cyclic rotations ([1,2,3,4,5], [2,3,4,5,1], ...) and 5 reverse cyclic rotations ([5,4,3,2,1], [4,3,2,1,5], ...). This design ensures that every score appears exactly twice in each position, effectively canceling out any systematic positional preference while preserving the semantic signal of the scores.

For the second rubric setting, we only need to replace the score rotations mentioned above with rotations of the different aspects. In addition, as a lower-cost alternative, we also experimented with directly randomizing the rubrics.

\section{Experimental Setup}
We evaluated position bias using 6 open-weight models across four datasets. Two of these datasets include human ratings, which allowed us to compute the correlation between LLM scores and human judgments under the permutation-based setting.

\subsection{Datasets}
\begin{table}[htbp]
  \centering
  \caption{Dataset Statistics}
  \resizebox{\columnwidth}{!}{%
  \begin{tabular}{lcr}
  \toprule
  \textbf{Dataset} & \textbf{Composition} & \textbf{Total} \\
  \midrule
  MT-Bench & 80 prompts $\times$ 4 models & 320 \\
  Vicuna-Bench & 80 prompts $\times$ 4 models & 320 \\
  HANNA & 96 stories $\times$ 6 criteria & 576 \\
  SummEval & 100 articles $\times$ 4 models $\times$ 4 criteria & 1,600 \\
  \midrule
  \textbf{Total} & & \textbf{2,816} \\
  \bottomrule
  \end{tabular}
  }
  \label{tab:dataset_stats}
\end{table}
We use MT-Bench, HANNA, SummEval, and Vicuna-Bench as our datasets for evaluation. Our data are either taken from~\citet{kim2024prometheus} or adapted following their format, and therefore differ from those used in the original paper. Detailed descriptions of the datasets are outlined below:
\begin{itemize}
    \item \textbf{MT Bench}~\cite{mtbench}: A multi-turn conversation dataset with 80 prompts across 8 categories (writing, roleplay, reasoning, math, coding, extraction, STEM, humanities). ChatGPT~\cite{openai2023gpt04}, Llama-2-Chat-13B~\cite{touvron2023llama}, Vicuna-13B~\cite{vicuna2023}, and WizardLM-13B~\cite{DBLP:conf/iclr/XuSZG0FTLJ24} were used to generate responses, resulting in a total of 320 responses.
    \item \textbf{Vicuna Bench}~\cite{vicuna2023}: A single-turn conversation benchmark with 80 prompts across 9 categories (generic, knowledge, roleplay, common-sense, fermi, counterfactual, coding, math, writing). Similar to MT Bench, it has 320 responses from 4 models.
    \item \textbf{HANNA}~\cite{hanna}: A story generation benchmark. We use a subset of 96 stories written by humans across 6 criteria (Coherence, Empathy, Surprise, Engagement, Relevance, Complexity). Each evaluation is annotated by 3 human raters.
    \item \textbf{SummEval}~\cite{summeval}: A summarization evaluation benchmark with 100 CNN/DailyMail articles, each summarized by 4 models on 4 criteria (coherence, consistency, fluency, relevance). Each evaluation is made by 3 human expert raters.
\end{itemize}

\subsection{Judge Models}
We evaluate six open-weight LLMs spanning three families and two size points: \textbf{GPT-OSS-20B} and \textbf{GPT-OSS-120B}~\cite{openai2025gpt0oss0120b}, \textbf{Qwen3.5-9B} and \textbf{Qwen3.5-27B}~\cite{qwen3.5}, and \textbf{Gemma-3-12B} and \textbf{Gemma-3-27B}~\cite{gemmateam2025gemma3technicalreport}. Inference uses vLLM~\cite{kwon2023efficient}. Unless stated, the default temperature is 0. We report a temperature sweep $\tau\in\{0.0,0.3,0.6,1.0\}$ for all the models and disable reasoning mode for Qwen3.5. The gpt-oss family supports an explicit reasoning-effort setting; we sweep $\{\text{low}, \text{medium}, \text{high}\}$ for both sizes. The details of the prompt are shown in~\Cref{app:prompt}.

\subsection{Evaluation Protocol}
\label{sec:evaluation-protocol}
\begin{figure}[htbp]
\centering
\includegraphics[scale=0.4]{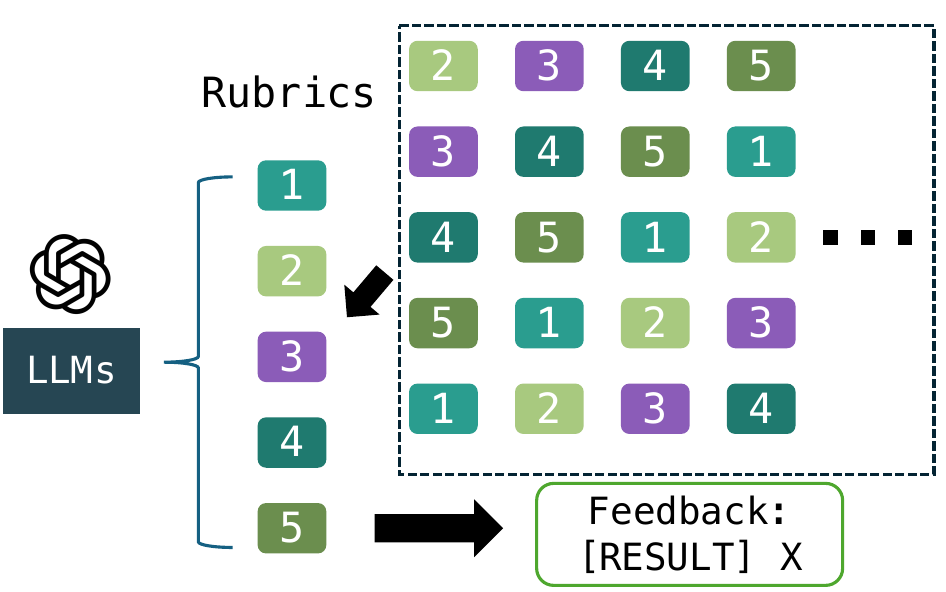}
\caption{An illustration of balanced permutations of rubric scores, ensuring that each score appears equally often at each position.}
\label{fig:flow}
\end{figure}

We study position bias along the two settings above (~\Cref{sec:balanced_permutation}), each with its own permutation, as illustrated in~\Cref{fig:flow}.

\paragraph{Score-option ordering (Setting 1).}
For the $n$-point rubric we permute the order of the score options and compare three strategies, each producing $K{=}10$ judgments per item: (i)~\textsc{Balanced}, the $2n$ cyclic rotations of~\Cref{sec:balanced_permutation}; (ii)~\textsc{Random}, $K$ orderings drawn at random; and (iii)~\textsc{Fixed}, a single canonical ordering $[1,\dots,n]$ repeated $K$ times. To see how the bias scales with rubric granularity we repeat the balanced experiment at $n\in\{2,3,5,9\}$, and to see how many orderings are actually needed we sub-sample $K\in\{1,\dots,10\}$ from the balanced and random sets.

\paragraph{Criterion ordering (Setting 2).}
Only HANNA (6 criteria) and SummEval (4 criteria) have a multi-criterion structure; MT-Bench and Vicuna provide a single overall score and therefore do not support this setting. When a prompt scores several criteria at once we permute the order of the criteria with the same balanced cyclic construction ($2n_c$ orderings; 12 for HANNA, 8 for SummEval), holding each criterion's score scale fixed, and compare three budget-matched strategies (\textsc{Balanced} / \textsc{Random} / \textsc{Fixed} criterion orderings, all at $2n_c$ reads per item), mirroring Setting 1.

\subsection{Metrics}
\paragraph{Bias detection.}
We test whether the selected-position distribution deviates from uniform with a $\chi^2$ goodness-of-fit test against $1/n$. To compare bias strength across rubric sizes we report Cram\'er's $V=\sqrt{\chi^2/(N(n-1))}\in[0,1]$, which is invariant to sample size. For criterion ordering we use an item-blocked Friedman test (blocking on the story/article).

\paragraph{Human alignment.}
We measure agreement with human ratings by Pearson's correlation coefficient $r$ and Spearman's rank correlation coefficient $\rho$, with 95\% bootstrap confidence intervals. To compare two strategies we use a paired bootstrap on the difference $\Delta r$ (resampling the same items for both correlations), which controls for the high item-level correlation between strategies.

\paragraph{Downstream impact.}
On the multi-candidate datasets we rank the candidate responses per prompt under each strategy and report the mean per-prompt Kendall $\tau$ between strategies and the fraction of prompts whose top-1 candidate changes.
 
\section{Results}

\subsection{Model-Specific Position Bias (Setting 1)}

\begin{figure*}[htbp]
\centering
\includegraphics[width=0.7\textwidth]{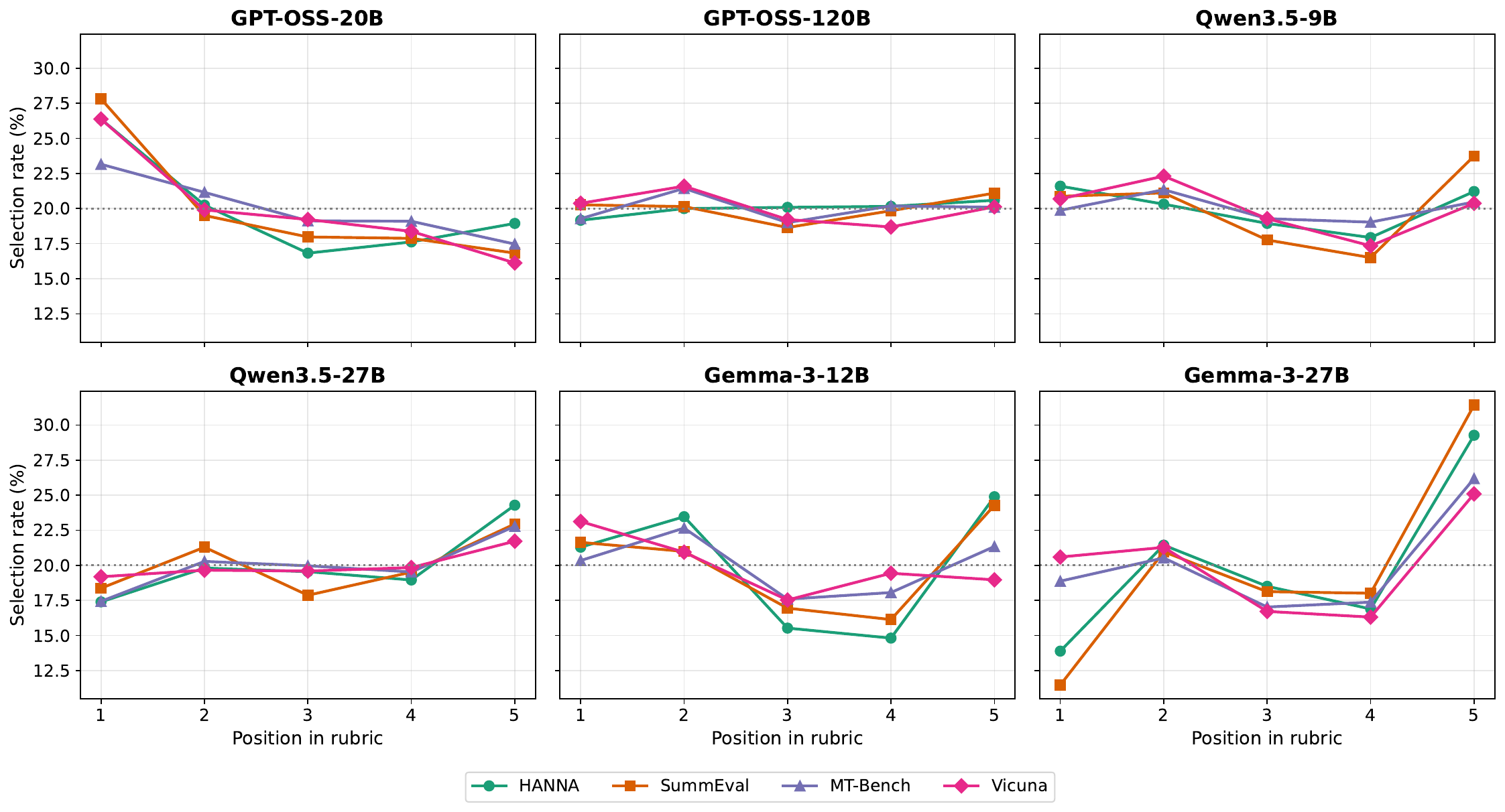}
\caption{Per-judge position-selection distribution under the 5-point rubric (one solid line per dataset, distinct color and marker), averaged over the 10 balanced orderings; the dotted line marks the 20\% uniform baseline. The direction of bias is model-specific yet consistent across datasets: GPT-OSS-20B over-selects Pos~1, Gemma-3-27B over-selects Pos~5, and GPT-OSS-120B is close to uniform.}
\label{fig:position_bias}
\end{figure*}

\autoref{fig:position_bias} shows the marginal position-selection distribution on the standard 5-point rubric, averaged over the 10 balanced orderings. If there is no position bias, every line would be near the 20\% baseline. \autoref{tab:scale_bias_5pt} reports the $\chi^2$ goodness-of-fit against uniform together with the Pos1 / Pos5 selection rates on the two datasets with human labels. Every $\chi^2$ on HANNA and SummEval is significant at $p<0.05$, the sole exception being the near-uniform GPT-OSS-120B on HANNA; on the smaller MT-Bench and Vicuna ($n{=}320$) the test has less power and a few cells fall short.

\begin{table*}[t]
\centering
\caption{5-point rubric: $\chi^2$ on position selection against uniform $1/5$, with Pos1 and Pos5 selection rates (\%), for all four datasets. Larger $\chi^2$ = stronger bias; the significance threshold is $\chi^2_{4,0.05}=9.49$. All cells on the human-rated HANNA and SummEval are significant except GPT-OSS-120B on HANNA; $^{\dagger}$ marks the cells that fall below significance (the low-power MT-Bench / Vicuna cells at $n{=}320$, plus GPT-OSS-120B / HANNA).}
\label{tab:scale_bias_5pt}
\small
\setlength{\tabcolsep}{3pt}
\begin{tabular}{l|rrr|rrr|rrr|rrr}
\toprule
\multirow{2}{*}{Model} & \multicolumn{3}{c|}{HANNA} & \multicolumn{3}{c|}{SummEval} & \multicolumn{3}{c|}{MT-Bench} & \multicolumn{3}{c}{Vicuna}\\
\cmidrule(lr){2-4}\cmidrule(lr){5-7}\cmidrule(lr){8-10}\cmidrule(lr){11-13}
& $\chi^2$ & Pos1 & Pos5 & $\chi^2$ & Pos1 & Pos5 & $\chi^2$ & Pos1 & Pos5 & $\chi^2$ & Pos1 & Pos5\\
\midrule
GPT-OSS-20B  &  165.7 & 26.4 & 18.9 &  639.8 & 27.8 & 16.8 &  30.9 & 23.2 & 17.5 &  94.3 & 26.4 & 16.1\\
GPT-OSS-120B &  3.1$^{\dagger}$ & 19.2 & 20.6 &   25.2 & 20.3 & 21.1 &   5.8$^{\dagger}$ & 19.3 & 20.1 &   8.0$^{\dagger}$ & 20.4 & 20.1\\
Qwen3.5-9B   &   27.4 & 21.6 & 21.2 &  265.9 & 20.9 & 23.7 &   5.6$^{\dagger}$ & 19.9 & 20.5 &  21.7 & 20.7 & 20.4\\
Qwen3.5-27B  &   76.3 & 17.4 & 24.3 &  142.5 & 18.4 & 22.9 &  23.4 & 17.4 & 22.8 &   6.3$^{\dagger}$ & 19.2 & 21.7\\
Gemma-3-12B  &  243.2 & 21.3 & 24.9 &  369.8 & 21.6 & 24.3 &  29.6 & 20.3 & 21.3 &  29.0 & 23.1 & 19.0\\
Gemma-3-27B  &  395.3 & 13.9 & 29.3 & 1691.7 & 11.5 & 31.4 &  88.9 & 18.9 & 26.2 &  83.7 & 20.6 & 25.1\\
\bottomrule
\end{tabular}
\end{table*}

Another finding is that the direction of bias differs across judge models. GPT-OSS-20B is more first-biased; Gemma-3-27B and Qwen3.5-27B are more last-biased; GPT-OSS-120B is near-uniform. So the bias here is a property of the model, not of the rubric layout, given that we use the same prompts.

\subsection{When Does Permutation Help? (Setting 1)}
The permutation operation can definitely reduce bias, because the scores are placed evenly across all positions. However, reducing this kind of bias does not necessarily lead to increased alignment with humans.

We compare three budget-matched strategies, each producing $K{=}10$ judgments per item: \textsc{Balanced} ($2n$ cyclic rotations), \textsc{Random} ($K$ random orderings), and \textsc{Fixed} (the canonical order repeated $K$ times). Because each item's score is the mean over $K$ reads, the \textsc{Balanced}$-$\textsc{Fixed} gap isolates de-biasing from variance reduction, while \textsc{Balanced}$-$\textsc{Random} isolates exact balance from mere aggregation. \autoref{fig:variance_ablation} shows the per-strategy Pearson $r$ with 95\% bootstrap CIs and \autoref{tab:variance_paired} the paired bootstrap CIs on the differences (resampling the same items for both correlations).

\begin{table*}[t]
\centering
\caption{Paired bootstrap 95\% CI for $\Delta r=r_{\text{Balanced}}-r_{\text{Random}}$ and $r_{\text{Balanced}}-r_{\text{Fixed}}$ (2000 iterations, same items resampled for both correlations). CIs excluding 0 are paired-significant at $p<0.05$. \textbf{Bold} marks paired-significant cells; \textsc{Balanced} significantly beats \textsc{Fixed} on 8 of 12 cells and no cell has \textsc{Fixed} beating \textsc{Balanced}.}
\label{tab:variance_paired}
\small
\begin{tabular}{ll|cc|cc}
\toprule
\multirow{2}{*}{Model} & \multirow{2}{*}{Dataset} & \multicolumn{2}{c|}{Balanced $-$ Random} & \multicolumn{2}{c}{Balanced $-$ Fixed} \\
\cmidrule(lr){3-4} \cmidrule(lr){5-6}
 & & $\Delta r$ & 95\% CI & $\Delta r$ & 95\% CI \\
\midrule
GPT-OSS-20B  & HANNA    & $+0.009$ & [$-$.013,$+$.031] & $\mathbf{+0.075}$ & $\mathbf{[+.038,+.116]}$ \\
GPT-OSS-20B  & SummEval & $+0.009$ & [$-$.003,$+$.021] & $\mathbf{+0.083}$ & $\mathbf{[+.057,+.108]}$ \\
GPT-OSS-120B & HANNA    & $\mathbf{+0.014}$ & $\mathbf{[+.000,+.029]}$ & $+0.021$ & [$-$.005,$+$.046] \\
GPT-OSS-120B & SummEval & $-0.002$ & [$-$.010,$+$.006] & $\mathbf{+0.018}$ & $\mathbf{[+.003,+.033]}$ \\
Qwen3.5-9B   & HANNA    & $-0.000$ & [$-$.016,$+$.017] & $-0.009$ & [$-$.047,$+$.029] \\
Qwen3.5-9B   & SummEval & $-0.008$ & [$-$.018,$+$.004] & $\mathbf{+0.050}$ & $\mathbf{[+.027,+.072]}$ \\
Qwen3.5-27B  & HANNA    & $-0.006$ & [$-$.021,$+$.011] & $\mathbf{+0.076}$ & $\mathbf{[+.047,+.105]}$ \\
Qwen3.5-27B  & SummEval & $+0.007$ & [$-$.001,$+$.014] & $+0.011$ & [$-$.003,$+$.026] \\
Gemma-3-12B  & HANNA    & $-0.006$ & [$-$.033,$+$.022] & $-0.004$ & [$-$.043,$+$.034] \\
Gemma-3-12B  & SummEval & $+0.001$ & [$-$.011,$+$.013] & $\mathbf{+0.035}$ & $\mathbf{[+.014,+.055]}$ \\
Gemma-3-27B  & HANNA    & $+0.009$ & [$-$.016,$+$.034] & $\mathbf{+0.070}$ & $\mathbf{[+.036,+.106]}$ \\
Gemma-3-27B  & SummEval & $\mathbf{+0.015}$ & $\mathbf{[+.001,+.028]}$ & $\mathbf{+0.027}$ & $\mathbf{[+.010,+.045]}$ \\
\bottomrule
\end{tabular}
\end{table*}

We have two findings. (i) \textbf{Balanced $\approx$ Random}: the paired CI for $r_{\text{Balanced}}-r_{\text{Random}}$ contains 0 on 10 of 12 cells (point estimates in $[-0.008,+0.015]$), so exact score--position balance buys little over random ordering at $K{=}10$. The improvement comes from aggregating distinct orderings, i.e.\ variance reduction. (ii) \textbf{Balanced $>$ Fixed only conditionally}: \textsc{Balanced} detectably beats \textsc{Fixed} on 8 of 12 cells, again the judges with the strongest position bias (GPT-OSS-20B, Qwen3.5-27B, Gemma-3-27B), and never loses to it; the remaining cells are inconclusive. In short, ``removing the bias'' is mechanical, but ``improving human correlation'' is conditional. Permutation helps human agreement only when the judge is strongly biased to begin with.

\subsection{How Many Orderings?}

\textsc{Balanced} needs exactly $2n$ orderings for exact balance, but how many are actually useful? We sub-sample $K\in\{1,\dots,10\}$ from both the balanced and the random sets, average each item over the chosen orderings, and plot mean Pearson $r$ vs $K$ (\autoref{fig:budget}). The random and balanced curves track each other within $0.02$ at every $K$ on 5 of 6 judges, and the gain saturates early: for judge models that benefit, roughly two-thirds of the $K{=}1\!\to\!10$ improvement is reached by $K{=}3$ and about $85\%$ by $K{=}5$.

\subsection{Bias Across Rubric Scales}

To rule out a 5-point artefact we repeat the balanced experiment at $n\in\{2,3,5,9\}$, selecting evenly spaced descriptions from the original rubric. Because $\chi^2$ scales with sample size, we compare bias strength with Cramér's $V=\sqrt{\chi^2/(N(n-1))}\in[0,1]$ (\autoref{tab:scale_summary}). Bias does not vanish at any scale, and it is not monotone: on 5 of 6 judge models the lowest $V$ sits at an intermediate scale ($n{=}3$ or $5$) while the extremes ($n{=}2$ and $9$) carry the highest $V$. Magnitude tracks the judge family, not size: GPT-OSS-120B stays $\leq0.05$ everywhere, whereas Gemma-3-27B is the most biased at every scale and roughly doubles from $n{=}5$ ($0.114$) to $n{=}9$ ($0.220$). Migrating to a finer 9-point or a coarser binary rubric therefore tends to increase bias; 3- or 5-point is the lower-bias regime.

\begin{table}[t]
\centering
\caption{Mean Cramér's $V$ for position-selection bias by judge and rubric resolution ($V\in[0,1]$, invariant to sample size; higher = stronger bias). Each cell averages over the datasets evaluated at that scale (4 for $n\in\{2,3,5\}$; HANNA + SummEval for $n{=}9$).}
\label{tab:scale_summary}
\small
\begin{tabular}{l|cccc}
\toprule
Model & $n{=}2$ & $n{=}3$ & $n{=}5$ & $n{=}9$ \\
\midrule
GPT-OSS-20B  & 0.035 & 0.035 & 0.080 & 0.036 \\
GPT-OSS-120B & 0.049 & 0.025 & 0.019 & 0.022 \\
Qwen3.5-9B   & 0.098 & 0.064 & 0.040 & 0.071 \\
Qwen3.5-27B  & 0.096 & 0.060 & 0.042 & 0.065 \\
Gemma-3-12B  & 0.168 & 0.059 & 0.069 & 0.158 \\
Gemma-3-27B  & 0.172 & 0.066 & 0.114 & 0.220 \\
\bottomrule
\end{tabular}
\end{table}

\subsection{Criterion-Order Bias (Setting 2)}

When a prompt scores several criteria at once (HANNA: 6, SummEval: 4), the order in which the criteria are listed is a second bias axis, orthogonal to score-option order. We permute the criterion order with the same balanced construction and test, per criterion, whether its score depends on its position with an item-blocked Friedman test (blocking on the story/article). \autoref{tab:criterion} reports the per-judge gap $\Delta$Pos between the best- and worst-position mean score and the number of significant criteria; \autoref{fig:criterion} visualises the per-criterion position effect.

\begin{table}[t]
\centering
\caption{Multi-criterion ordering. ``Avg $r$'' is mean Pearson with humans across criteria; ``Mean / Max $\Delta$Pos'' is the best$-$worst-position score gap (5-point scale) averaged / maxed over criteria; ``Sig.'' counts criteria with a significant item-blocked Friedman test ($p<0.05$) out of the total. H represents HANNA; S represents SummEval.}
\label{tab:criterion}
\small
\begin{tabular}{l|r|rr|r}
\toprule
Model & Avg $r$ & Mean $\Delta$ & Max $\Delta$ & Sig. \\
\midrule
GPT-OSS-20B(H) & 0.290 & 0.26 & 0.39 & 6/6 \\
GPT-OSS-20B(S) & 0.433 & 0.23 & 0.39 & 4/4 \\
GPT-OSS-120B(H) & 0.323 & 0.25 & 0.54 & 5/6 \\
GPT-OSS-120B(S) & 0.421 & 0.30 & 0.46 & 4/4 \\
Qwen3.5-9B(H) & 0.263 & 0.45 & 0.77 & 6/6 \\
Qwen3.5-9B(S) & 0.444 & 0.53 & 0.80 & 4/4 \\
Qwen3.5-27B(H) & 0.337 & 0.33 & 0.43 & 6/6 \\
Qwen3.5-27B(S) & 0.451 & 0.39 & 0.79 & 4/4 \\
Gemma-3-12B(H) & 0.381 & 0.30 & 0.68 & 5/6 \\
Gemma-3-12B(S) & 0.486 & 0.21 & 0.29 & 4/4 \\
Gemma-3-27B(H) & 0.354 & 0.30 & 0.44 & 6/6 \\
Gemma-3-27B(S) & 0.507 & 0.43 & 0.61 & 4/4 \\
\bottomrule
\end{tabular}
\end{table}

The bias is pervasive: $58$ of $60$ (judge, criterion) Friedman tests are significant (all 24 SummEval cells, $34/36$ HANNA cells), and the most extreme cell (Qwen3.5-9B on SummEval) shifts a criterion's mean by up to $0.80$ points. A judge corrected for score-position bias can therefore still be biased through criterion order.

We then ask whether mitigating it helps, with the same budget-matched three-way ablation (\textsc{Balanced}/\textsc{Random}/\textsc{Fixed} criterion order, $2n_c$ reads: 12 for HANNA, 8 for SummEval). \autoref{tab:criterion_mitigation} reports macro-average Pearson with cluster-level paired-bootstrap CIs. The story mirrors the score axis exactly: budget-matched \textsc{Balanced}$-$\textsc{Fixed} is significant on 6 of 12 cells, and the \textsc{Balanced} and \textsc{Random} columns differ by at most $0.02$ on every cell.

\begin{table*}[t]
\centering
\caption{Budget-matched criterion-order ablation on HANNA (H) and SummEval (S). \textsc{Bal}/\textsc{Rnd}/\textsc{Fix} are macro-average Pearson with humans under Balanced/Random/Fixed criterion ordering (same $2n_c$-read budget); $\Delta$ is the cluster-level paired-bootstrap difference \textsc{Bal}$-$\textsc{Fix} (resampling stories / article--model pairs), \textbf{bold} when its $95\%$ CI excludes 0. De-biasing (\textsc{Bal}$-$\textsc{Fix}) is significant on 6/12 cells (three Gemma, both GPT-OSS SummEval, and GPT-OSS-20B HANNA). \textsc{Balanced} and \textsc{Random} differ by at most $0.02$ on every cell.}
\label{tab:criterion_mitigation}
\small
\setlength{\tabcolsep}{4pt}
\begin{tabular}{l|rrr|r|rrr|r}
\toprule
\multirow{2}{*}{Model} & \multicolumn{4}{c|}{HANNA (H)} & \multicolumn{4}{c}{SummEval (S)}\\
\cmidrule(lr){2-5}\cmidrule(lr){6-9}
 & Bal & Rnd & Fix & $\Delta$ & Bal & Rnd & Fix & $\Delta$\\
\midrule
GPT-OSS-20B  & 0.290 & 0.283 & 0.201 & $\mathbf{+0.090}$ & 0.433 & 0.434 & 0.399 & $\mathbf{+0.034}$\\
GPT-OSS-120B & 0.323 & 0.335 & 0.300 & $+0.022$ & 0.421 & 0.418 & 0.400 & $\mathbf{+0.021}$\\
Qwen3.5-9B   & 0.263 & 0.259 & 0.244 & $+0.019$ & 0.444 & 0.428 & 0.423 & $+0.021$\\
Qwen3.5-27B  & 0.337 & 0.337 & 0.330 & $+0.007$ & 0.451 & 0.445 & 0.445 & $+0.006$\\
Gemma-3-12B  & 0.381 & 0.393 & 0.338 & $\mathbf{+0.043}$ & 0.486 & 0.481 & 0.455 & $\mathbf{+0.031}$\\
Gemma-3-27B  & 0.354 & 0.343 & 0.347 & $+0.007$ & 0.507 & 0.504 & 0.483 & $\mathbf{+0.023}$\\
\bottomrule
\end{tabular}
\end{table*}

\subsection{Downstream Rank Reversal}

Finally we show the bias propagates into ranking, which best-of-$N$ selection, RLAIF reward modeling, and leaderboards all consume. On the multi-candidate datasets (MT-Bench, Vicuna, SummEval) we rank the candidate responses per prompt under \textsc{Balanced} and \textsc{Fixed} and compare (\autoref{tab:rank_reversal}). Across all 18 cells the per-prompt Kendall $\tau$ between the two rankings is only $0.65$--$0.85$, and the top-1 candidate flips on $15$--$41\%$ of prompts. This is not confined to high-bias judges: GPT-OSS-120B, the lowest-bias judge by $\chi^2$, still shows $21$--$38\%$ top-1 reversal, because rank reversal depends on per-item variance, not just the average direction of bias. A reward model fed these scores would see a different winning candidate on roughly a quarter of prompts from a nominally cosmetic ordering choice alone.

\begin{table}[t]
\centering
\caption{Rank reversal under ordering perturbation (\textsc{Balanced} vs \textsc{Fixed}): per (judge, dataset) we rank the candidate responses per prompt and report the mean per-prompt Kendall $\tau$ and the percentage of prompts whose top-1 candidate flips (t1). Datasets: M\,=\,MT-Bench, V\,=\,Vicuna ($80$ prompts each), S\,=\,SummEval ($400$); all have 4 candidates per prompt. HANNA is omitted: it has a single text per item, hence no competing candidates to rank. $\tau$ never exceeds $0.85$ and t1 reaches $15$--$41\%$.}
\label{tab:rank_reversal}
\small
\setlength{\tabcolsep}{3.5pt}
\begin{tabular}{l|cc|cc|cc}
\toprule
\multirow{2}{*}{Model} & \multicolumn{2}{c|}{M} & \multicolumn{2}{c|}{V} & \multicolumn{2}{c}{S}\\
\cmidrule(lr){2-3}\cmidrule(lr){4-5}\cmidrule(lr){6-7}
 & $\tau$ & t1 & $\tau$ & t1 & $\tau$ & t1\\
\midrule
GPT-OSS-20B  & .72 & 15.0 & .65 & 28.8 & .67 & 41.0\\
GPT-OSS-120B & .74 & 21.2 & .77 & 36.2 & .72 & 38.2\\
Qwen3.5-9B   & .76 & 28.8 & .79 & 16.2 & .73 & 38.0\\
Qwen3.5-27B  & .84 & 23.8 & .85 & 16.2 & .85 & 26.0\\
Gemma-3-12B  & .75 & 18.8 & .75 & 30.0 & .78 & 36.0\\
Gemma-3-27B  & .77 & 17.5 & .82 & 25.0 & .81 & 33.2\\
\bottomrule
\end{tabular}
\end{table}

\section{Discussion}
\paragraph{Detection versus correction.} The sources of bias are relatively diverse; even after position bias has been eliminated, other forms of bias still remain. Moreover, even human ratings are subject to a certain degree of bias, and therefore the similarity is not necessarily improved.

\paragraph{Impact on selection, not absolute scores.} Although the correlation does not necessarily show a substantial improvement, when we swap the order of the criteria, the resulting variation in scores exerts a considerable impact on the ranking. Therefore, applications of rubric-based LLM-as-a-Judge should pay attention to such scenarios. It is also worthwhile to assess the influence of this kind of noise on downstream applications, such as model training.

\paragraph{Practical guidance.} Even given the same prompt, different models exhibit different biases: some prefer the first position, some prefer the last position, and some are relatively balanced. Furthermore, the biases observed under two different settings also differ. In addition, although permutation works well in theory, if the goal is merely to improve correlation, randomly shuffling the order is sufficient.

\section*{Limitations}
Owing to budget constraints, our experiments were not conducted on the most recent closed-source LLMs. For the same reason, we did not attempt to apply our method directly to works that leverage rubrics for model training.

\section*{Acknowledgments}
This work was supported by JST Moonshot R\&D Program Grant Number JPMJMS2236 and JSPS KAKENHI Grant Number JP25H01149.

\bibliography{reference}

\appendix
\section{Potential Risk}
Our work has potential applications in detecting whether questionnaires have been completed by LLMs, and conversely, it could also be used to manipulate questionnaires when respondents rely on LLMs to answer. By arranging the options in a particular order, one may bias the LLM toward producing certain responses more frequently.

\section{LLM Instructions}
\label{app:prompt}
\begin{figure}[htbp]
\centering
\includegraphics[scale=0.4]{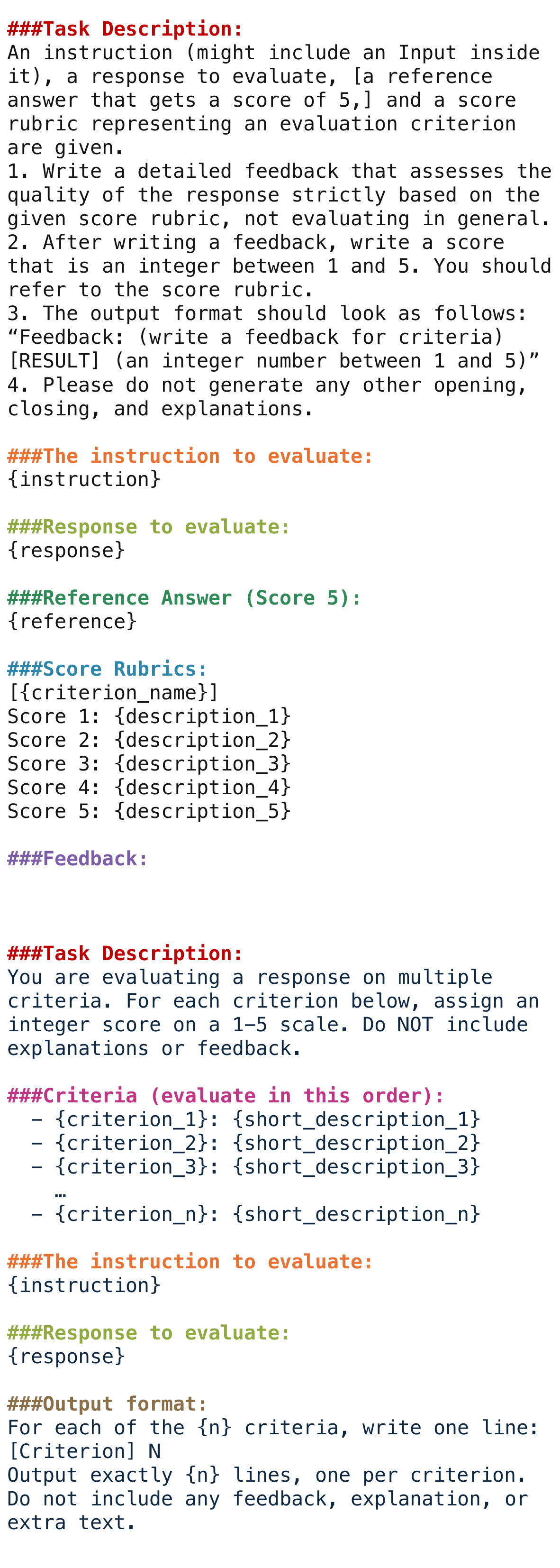}
\caption{Unified prompt format used in all experiments (Prometheus-Eval).}
\label{fig:prometheus_prompt}
\end{figure}

\Cref{fig:prometheus_prompt} illustrates the prompt template used in our experiments, which basically follows the style in~\citet{kim2024prometheus}. The LLM is provided with a task description, the instruction to evaluate, the response, a reference answer, and line-by-line score rubrics, and is asked to produce written feedback and a score from 1 to 5. Note that the HANNA and SummEval datasets do not include a reference answer in the prompt. For setting 2, we directly presented each criterion along with its description, and then asked the LLM to assign a score for each criterion.

\section{Sensitivity: Temperature and Reasoning Effort}
\begin{figure*}[htbp]
\centering
\includegraphics[width=1.0\textwidth]{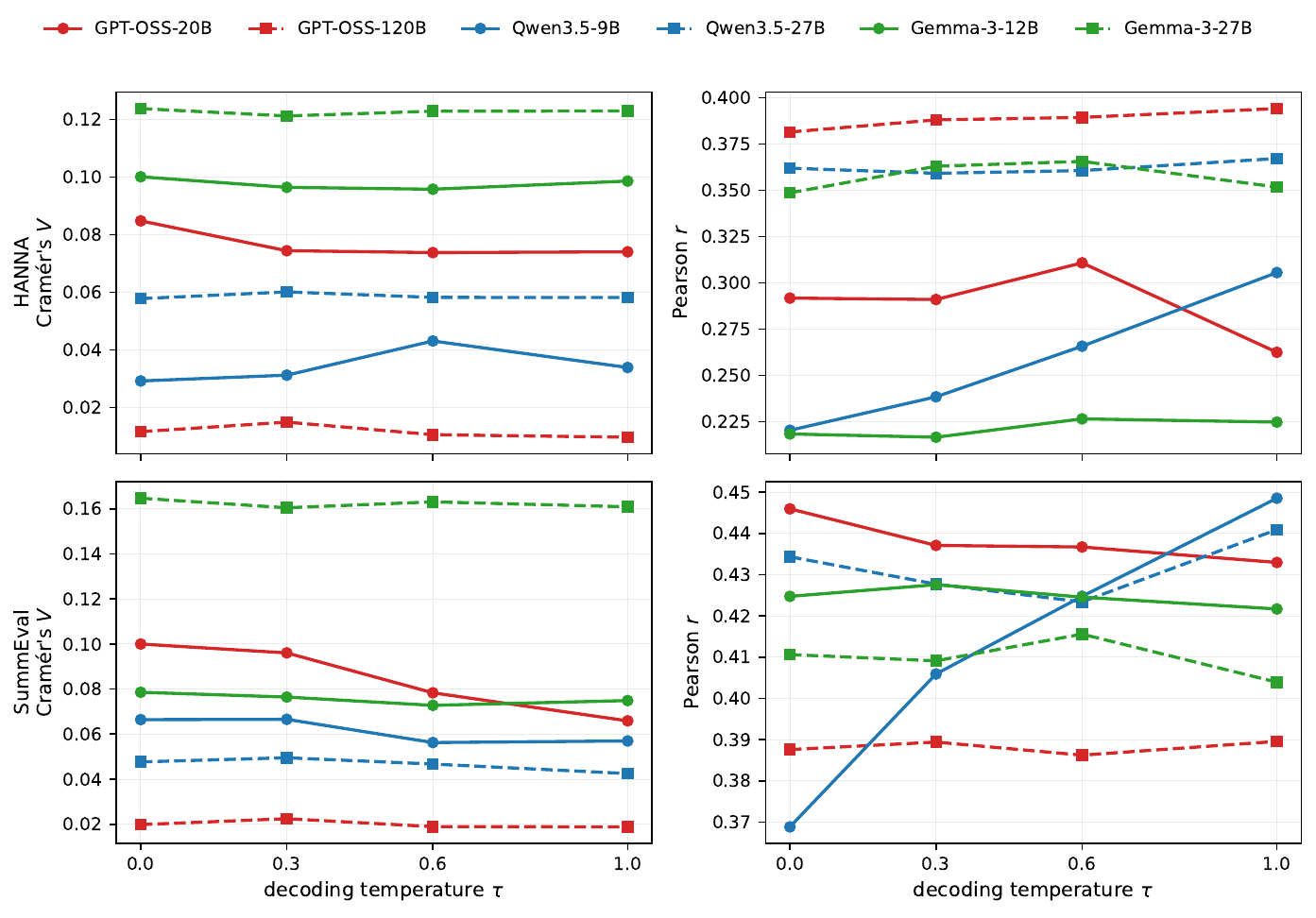}
\caption{Temperature sweep, all judges. Columns: Cramér's $V$ (left) and Pearson $r$ (right); rows: HANNA (top), SummEval (bottom); one solid line per judge. Bias is flat in $\tau$; only Qwen3.5-9B's $r$ rises.}
\label{fig:temp}
\end{figure*}

\begin{figure*}[htbp]
\centering
\includegraphics[width=1.0\textwidth]{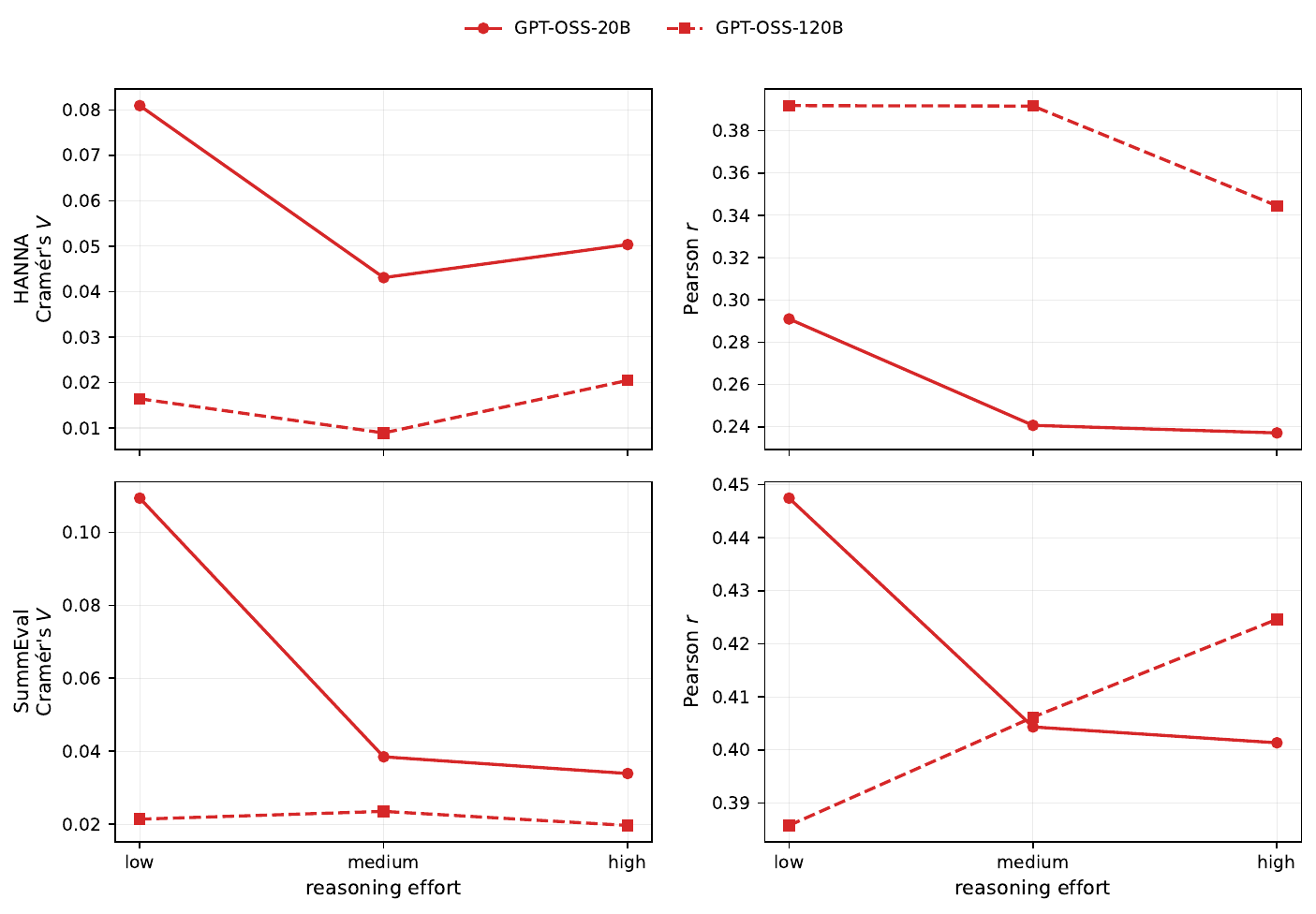}
\caption{Reasoning-effort sweep (gpt-oss). Same layout as \autoref{fig:temp}. Bias is non-monotone in effort and more effort does not consistently improve $r$.}
\label{fig:effort}
\end{figure*}

Neither temperature nor reasoning effort eliminates the bias. For decoding \textbf{temperature} (\autoref{fig:temp}), position bias ($\chi^2$ / Cramér's $V$) is flat across $\tau\in\{0,0.3,0.6,1.0\}$ for all six judges, and human $r$ is flat too, except Qwen3.5-9B, whose $r$ rises ($0.22\!\to\!0.31$ on HANNA) not by reducing bias but because sampling and averaging recover graded score resolution that greedy decoding discards. For \textbf{reasoning effort} (gpt-oss; \autoref{fig:effort}), bias is non-monotone. GPT-OSS-20B's HANNA $\chi^2$ drops $151\!\to\!43$ from low to medium then rises to $58$ at high. More effort does not reliably improve correlation. Both effects are model-specific.

\section{Position-Prior Decomposition}
\label{app:decomp}
We fit a conditional logit $u_{i\ell s}=\gamma_{i\ell}+\beta_\ell+\alpha_s$, splitting the score into content $\gamma_{i\ell}$, a leniency prior $\beta_\ell$, and a position prior $\alpha_s$. A likelihood-ratio test (LRT), which compares this fit against the case of forcing $\alpha=0$, rejects $\alpha=0$ on $23$ of $24$ cells, except GPT-OSS-120B on HANNA. We summarise the strength of the prior by $\mathrm{sd}(\alpha)$, the spread of $\alpha$ across the score positions (\autoref{tab:sdalpha}), where a larger value means a stronger bias. So, the bias is a real and estimable effect.

\begin{table}[t]
\centering
\caption{Position-prior spread $\mathrm{sd}(\alpha)$ from the conditional-logit fit on the 5-point rubric, averaged over the four datasets. A higher value means a stronger position prior.}
\label{tab:sdalpha}
\small
\begin{tabular}{l|c}
\toprule
Model & $\mathrm{sd}(\alpha)$ \\
\midrule
GPT-OSS-20B  & 0.40 \\
GPT-OSS-120B & 0.15 \\
Qwen3.5-9B   & 0.23 \\
Qwen3.5-27B  & 0.26 \\
Gemma-3-12B  & 0.52 \\
Gemma-3-27B  & 0.72 \\
\bottomrule
\end{tabular}
\end{table}

\section{Prompt Layout and Prompt-Level Mitigations}
\label{app:prompt_ablation}
We re-estimate $\alpha$ (\Cref{app:decomp}) using six prompts on the balanced 5-point design: \textsc{base} (the \texttt{Score N:} list), \textsc{prose} (same content, no list), \textsc{debias} (ignore the option order), \textsc{cot} (step-by-step check), \textsc{anchor} (two worked reference points at both ends), and \textsc{verify} (re-score after seeing its own canonical-order score). No setting removes the bias (\autoref{tab:prompt_arms}). \textsc{prose} even raises it, so the layout is not the reason. \textsc{anchor} gives the lowest value, which might be a good method to reduce the bias.

\begin{table}[t]
\centering
\caption{Position-prior spread $\mathrm{sd}(\alpha)$ by prompt arms, averaged over the $12$ (judge, dataset) cells (5-point balanced design). ``Rel.'' is the setting's mean $\mathrm{sd}(\alpha)$ relative to \textsc{base}; ``Sig.'' counts cells with a significant $\alpha$ (LRT $p<0.05$).}
\label{tab:prompt_arms}
\small
\begin{tabular}{l|ccc}
\toprule
Arm & mean $\mathrm{sd}(\alpha)$ & Rel.\ base & Sig. \\
\midrule
\textsc{base}   & 0.377 & --- & 11/12 \\
\textsc{prose}  & 0.455 & $1.21\times$ & 12/12 \\
\textsc{debias} & 0.373 & $0.99\times$ & 12/12 \\
\textsc{cot}    & 0.347 & $0.92\times$ & 11/12 \\
\textsc{verify} & 0.356 & $0.94\times$ & 12/12 \\
\textsc{anchor} & 0.249 & $0.66\times$ & 11/12 \\
\bottomrule
\end{tabular}
\end{table}

\section{Why Inference-Time Fixes Fail, and How Many Orderings Suffice}
\label{app:cost}
We measure how much of \textsc{Balanced}$\times10$'s correlation gain each rotation budget recovers~(\autoref{tab:cost}). A single fixed order is the baseline; repeating it ten times recovers only $8\%$ (greedy decoding is nearly deterministic, so the repeated runs add little). Averaging over different orderings helps more, but saturates fast: \textsc{Balanced}-$3$ already captures $70\%$ and \textsc{Balanced}-$5$ captures $87\%$.

\begin{table}[t]
\centering
\caption{Share of \textsc{Balanced}$\times10$'s human-correlation gain recovered by each strategy under greedy decoding, pooled over the human-labelled HANNA and SummEval cells. Only averaging over different orderings recovers most of the gain; repeating a single fixed order recovers almost none.}
\label{tab:cost}
\small
\begin{tabular}{l|c|c}
\toprule
Strategy & Cost & Gain captured \\
\midrule
\textsc{Fixed} (single read)            & $\times1$   & $0\%$ (baseline) \\
\textsc{Fixed}$\times10$ (repeated read) & $\times10$  & $8\%$ \\
\textsc{Balanced}-3                      & $\times3$   & $70\%$ \\
\textsc{Balanced}-5                      & $\times5$   & $87\%$ \\
\textsc{Balanced}-10                     & $\times10$  & $100\%$ \\
\bottomrule
\end{tabular}
\end{table}

\section{Additional Figures}
This appendix collects supporting figures referenced from the main text; the corresponding numbers appear in the tables there.
\subsection{Strategy Ablation (Forest Plots)}
\begin{figure*}[htbp]
\centering
\includegraphics[width=0.9\textwidth]{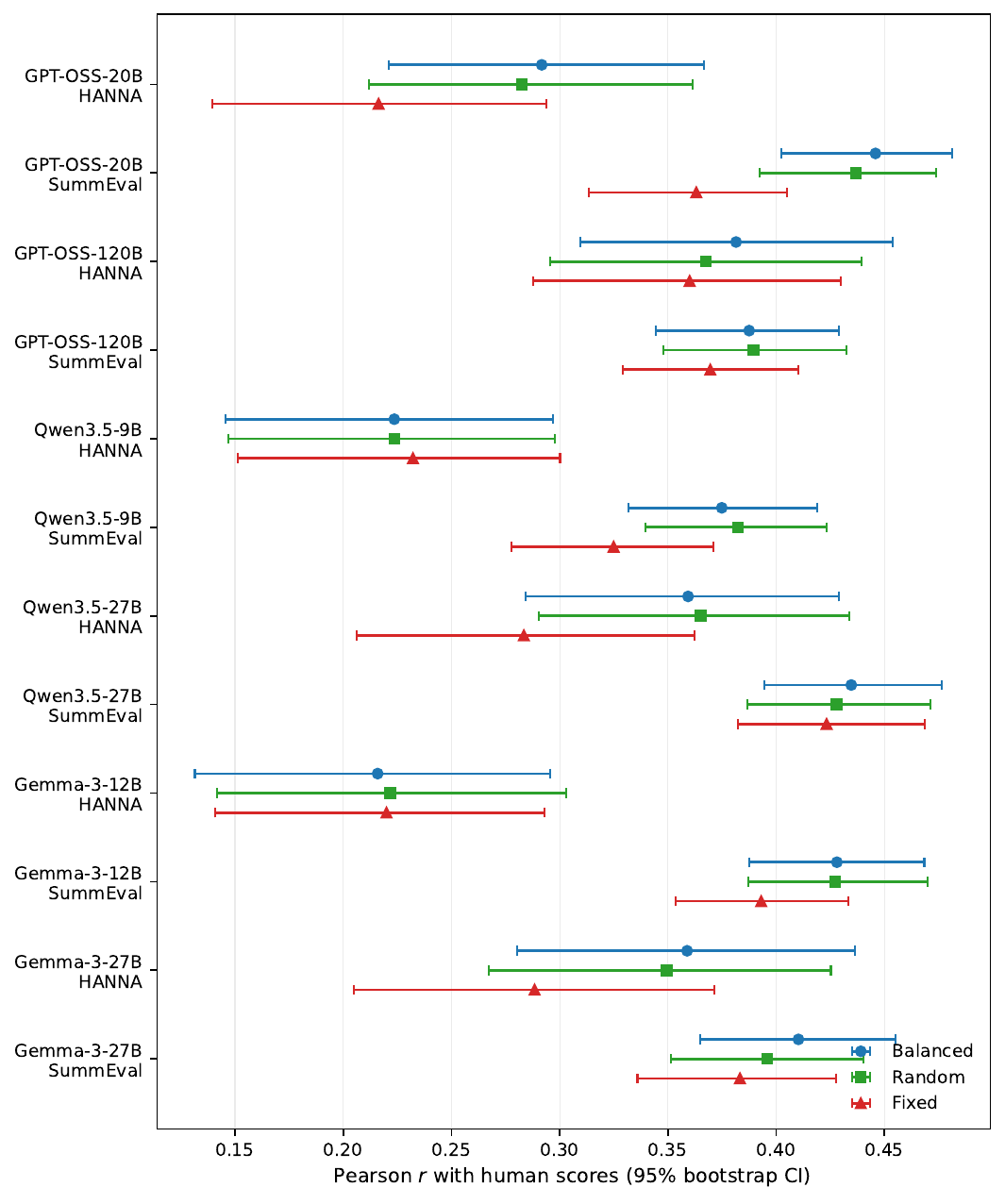}
\caption{Forest plot of Pearson $r$ (95\% bootstrap CI) for the three ordering strategies on HANNA and SummEval. \textsc{Balanced} and \textsc{Random} CIs are nearly identical on every row; \textsc{Fixed} is clearly below both only on the highest-bias judges.}
\label{fig:variance_ablation}
\end{figure*}

\autoref{fig:variance_ablation} is the visual companion to the paired comparison in \autoref{tab:variance_paired}: for each (judge, dataset) it plots the Pearson $r$ and its 95\% bootstrap CI under \textsc{Balanced}, \textsc{Random}, and \textsc{Fixed} score-option orderings, all at the matched $K{=}10$ budget. The \textsc{Balanced} and \textsc{Random} intervals are nearly identical on every row, confirming that exact balance buys little over random aggregation; \textsc{Fixed} drops clearly below both only on the strongly biased judges.

\subsection{Budget Sweep}
\begin{figure*}[htbp]
\centering
\includegraphics[width=0.95\textwidth]{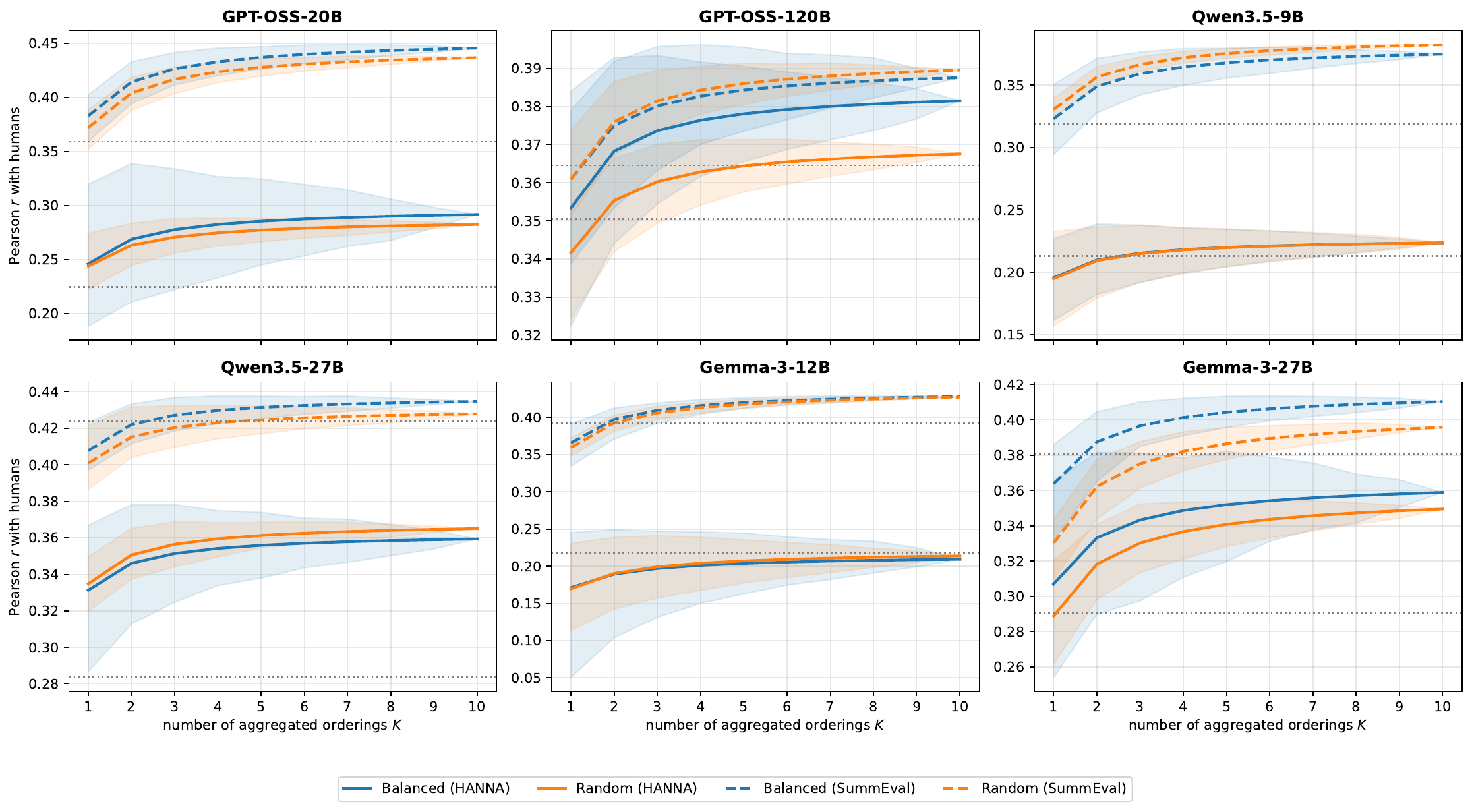}
\caption{Correlation with humans vs the number of aggregated orderings $K$, for \textsc{Random} and \textsc{Balanced} sub-sampling on HANNA (solid) and SummEval (dashed); the \textsc{Fixed} baseline is the dotted line. Bands are 2.5--97.5 percentiles over subsets. The two curves are nearly identical and most of the gain is realised by $K\!\approx\!3$.}
\label{fig:budget}
\end{figure*}

\autoref{fig:budget} supports the budget analysis in the main text. For each budget $K\in\{1,\dots,10\}$ we enumerate every size-$K$ subset of the $10$ orderings, average each item's score over the chosen orderings, and compute Pearson $r$ with humans (complete-case items); the curve is the mean over subsets and the band its $2.5$--$97.5$ percentile range. \textsc{Random} and \textsc{Balanced} sub-sampling track each other at every $K$, and most of the achievable gain is reached by $K\!\approx\!3$, so a handful of randomly ordered variants suffices.

\subsection{Criterion-Order Heatmaps}
\begin{figure*}[htbp]
\centering
\includegraphics[width=0.95\textwidth]{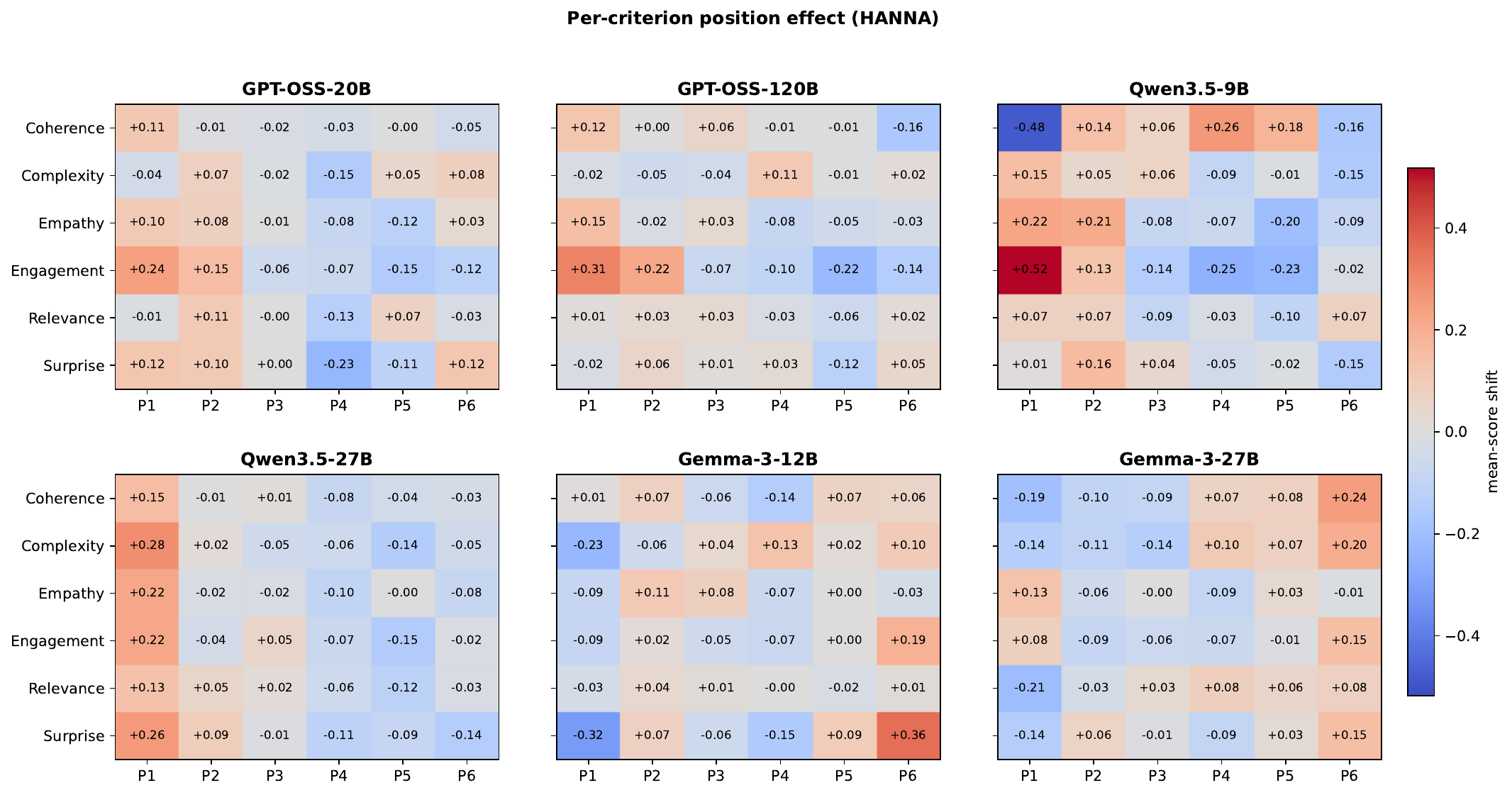}
\caption{Per-criterion position effect on HANNA for all six judges. Each cell is the shift in a criterion's mean score (rows) when placed at a given position (columns) relative to its overall mean; warm = higher there, cool = lower. Patterns are model-specific (e.g.\ Qwen3.5-9B scores Engagement $+0.52$ at Pos~1 but $-0.25$ at Pos~4). The SummEval counterpart is \autoref{fig:criterion_summeval}.}
\label{fig:criterion}
\end{figure*}

\autoref{fig:criterion} is the visual form of the criterion-order detection in \autoref{tab:criterion}: it shows, per judge, how much each criterion's mean score shifts with its position in the prompt. The off-diagonal structure makes the criterion-order bias directly visible, and its pattern is model-specific.

\begin{figure*}[htbp]
\centering
\includegraphics[width=0.95\textwidth]{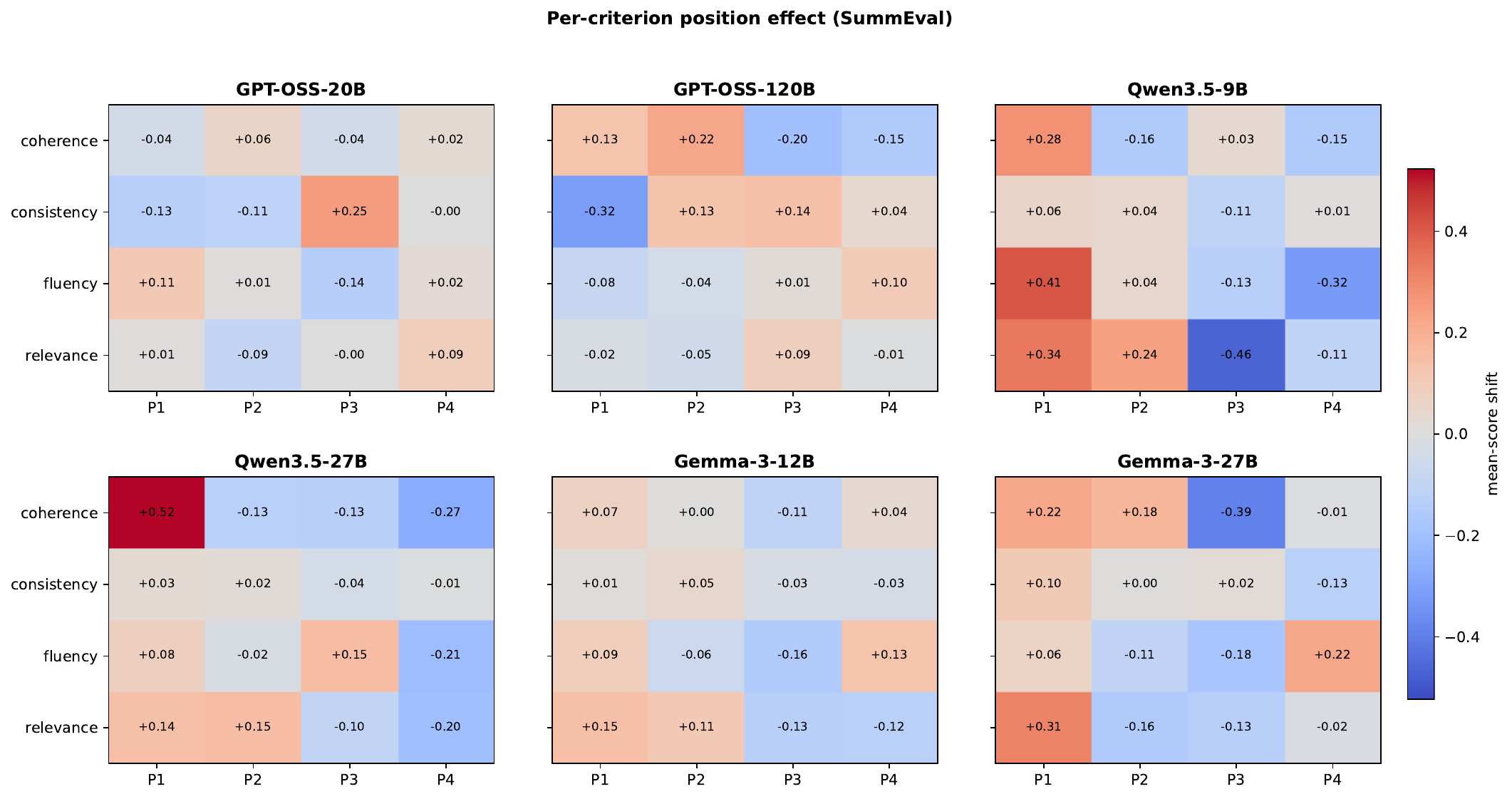}
\caption{Per-criterion position effect on SummEval (4 criteria) for all six judges; same construction as \autoref{fig:criterion}. The criterion-order effect shows the same qualitative, model-specific pattern as on HANNA.}
\label{fig:criterion_summeval}
\end{figure*}

\section{Licenses and Intended Use}
\label{app:licenses}
We use only publicly released benchmarks and open-weight models, all for research-purpose evaluation, consistent with their intended use; we redistribute no data or model weights.
\paragraph{Datasets.} MT-Bench and Vicuna-Bench are distributed with the FastChat project (Apache-2.0); SummEval and HANNA are released under the MIT license. Our prompts follow the Prometheus-Eval format~\cite{kim2024prometheus} and are adapted from these sources.
\paragraph{Models.} GPT-OSS-20B/120B and the Qwen3.5 family are released under Apache-2.0; the Gemma-3 family is released under the Gemma Terms of Use. All datasets and models are used in accordance with their licenses and intended purpose (benchmarking and evaluation research). The benchmarks contain no personally identifying information beyond what their original public releases already include.

\section{Computational Budget and Infrastructure}
\label{app:compute}
All experiments are inference-only; we train no models. Part of the results was obtained through the OpenRouter API; the remaining judges are served locally with vLLM~0.20.1 on a single NVIDIA H100 (96\,GB) and queried on the same machine. The full study comprises roughly $2.1$M judge calls across the six models and all conditions (the four datasets, scales $n\in\{2,3,5,9\}$, the temperature and reasoning-effort sweeps, and the three permutation strategies). The local GPU inference budget is under $50$ GPU-hours.

\section{Statistical Tools}
\label{app:stats}
All statistical tests use \texttt{scipy.stats} (v1.17.1): the $\chi^2$ goodness-of-fit test (\texttt{chisquare}), the item-blocked Friedman test (\texttt{friedmanchisquare}), and Pearson / Spearman correlations (\texttt{pearsonr}, \texttt{spearmanr}). Cram\'er's $V$, the bootstrap confidence intervals, and the paired bootstrap on $\Delta r$ are our own implementations on top of \texttt{numpy} (v2.4.3): the $95\%$ confidence intervals use $1{,}000$ resamples and the paired bootstrap on $\Delta r$ uses $2{,}000$ resamples, all with a fixed random seed.

\section{Use of AI Assistants}
\label{app:ai}
We used an AI coding assistant to help write the experiment-running scripts and the analysis code. The code for generating figures and tables is also written with the help of an AI assistant. AI assistants also helped refine the manuscript.

\end{document}